\documentclass[runningheads]{llncs}
\usepackage{cite}
\usepackage{enumitem}
\usepackage{graphicx}

\usepackage{graphicx}
\usepackage{multirow}
\usepackage{amssymb}
\usepackage{amsmath}
\usepackage{booktabs}
\usepackage{pifont}
\usepackage[misc]{ifsym} 
\usepackage[table]{xcolor}
\usepackage[colorlinks,bookmarksopen,bookmarksnumbered,citecolor=blue, linkcolor=blue, urlcolor=blue]{hyperref}
\begin{document}
\title{Multi-Branch Auxiliary Fusion YOLO with Re-parameterization Heterogeneous Convolutional for accurate object detection}
%
\titlerunning{MAF-YOLO with RHConv for accurate object detection}
%
\author{Zhiqiang Yang\inst{1}, Qiu Guan\inst{1}\textsuperscript{(\Letter)}, Keer Zhao\inst{1}, Jianmin Yang\inst{2}, Xinli Xu\inst{1}, Haixia Long\inst{1}, Ying Tang\inst{1}\textsuperscript{(\Letter)}}
%
\authorrunning{Z. Yang et al.}
%
\institute{\textsuperscript{1}\makebox[0pt][]{1}Zhejiang University of Technology, Hangzhou, 310023, China\\
         gq@zjut.edu.cn\\
         \textsuperscript{2}\makebox[0pt][]{1}Zhejiang College of Sports, Hangzhou, 310012, China\\
         }
%
\maketitle              
\begin{abstract}
	Due to the effective performance of multi-scale feature fusion, Path Aggregation FPN (PAFPN) is widely employed in YOLO detectors. However, it cannot efficiently and adaptively integrate high-level semantic information with low-level spatial information simultaneously. We propose a new model named MAF-YOLO in this paper, which is a novel object detection framework with a versatile neck named Multi-Branch Auxiliary FPN (MAFPN). Within MAFPN, the Superficial Assisted Fusion (SAF) module is designed to combine the output of the backbone with the neck, preserving an optimal level of shallow information to facilitate subsequent learning. Meanwhile, the Advanced Assisted Fusion (AAF) module deeply embedded within the neck conveys a more diverse range of gradient information to the output layer.
    Furthermore, our proposed Re-parameterized Heterogeneous Efficient Layer Aggregation Network (RepHELAN) module ensures that both the overall model architecture and convolutional design embrace the utilization of heterogeneous large convolution kernels. Therefore, this guarantees the preservation of information related to small targets while simultaneously achieving the multi-scale receptive field. Finally, taking the nano version of MAF-YOLO for example, it can achieve 42.4\% AP on COCO with only 3.76M learnable parameters and 10.51G FLOPs, and approximately outperforms YOLOv8n by about 5.1\%. The source code of this work is available at: \href{https://github.com/yang-0201/MAF-YOLO}{https://github.com/yang-0201/MAF-YOLO}. 
\end{abstract}
\section{Introduction}
\begin{figure}[htb]
	
	\begin{minipage}[b]{1.0\linewidth}
		\centering
		\centerline{\includegraphics[width=12cm]{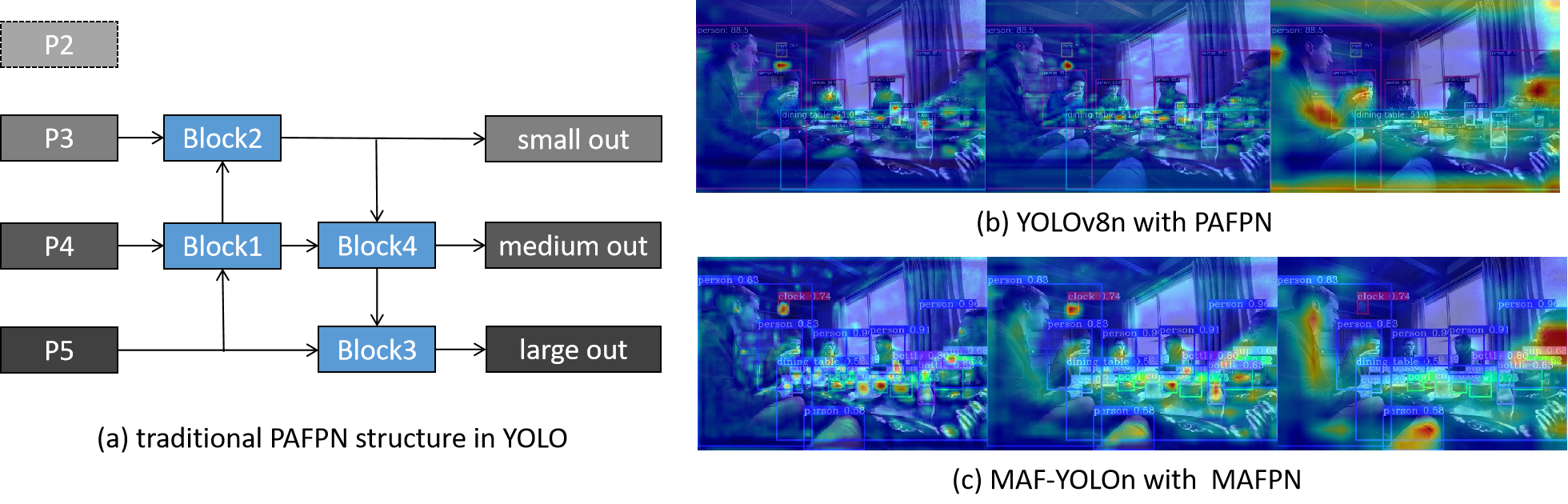}}
	\end{minipage}
	\caption{(a) represents the traditional PAFPN structure, (b) and (c) display the GradCAM++ visualization results of the neck for YOLOv8n and MAF-YOLOn. These three images represent the output layers of the model for small, medium, and large objects.}
	\label{mafpn}
\end{figure}

To implement real-time object detection with high performance, a variety of algorithms have been developed in recent years. Among them, a series of YOLO algorithms~\cite{yolo1,yolov3,yolov4,yolov5,yolox,yolov6,yolov7,yolov8,damo,rtmdet,ppyoloe,goldyolo,yolov9}, from YOLOv1 to YOLOv9, have played increasingly significant roles due to their compromise between speed and accuracy. However, a common shortcoming of YOLO series algorithms is the limitation of multi-scale feature fusion. Although the feature fusion mechanism of the Path Aggregation Feature Pyramid Network (PAFPN)~\cite{pafpn}, an improvement over the Feature Pyramid Network (FPN)~\cite{fpn}, has been widely integrated into YOLO. This mechanism introduces a dual-path approach to enhance feature integration, thereby improving accuracy while also controlling computational costs.
In Fig.~\ref{mafpn}(a), P3, P4, and P5 represent the output information of different levels of the backbone. The neck structure of the YOLO series utilizes a traditional PAFPN, which incorporates two main paths for multi-scale feature fusion. Nevertheless, PAFPN still possesses two significant limitations.

Firstly, PAFPN tends to merge homogenous scale feature maps and lacks integrated processing and fusion of multi-scale information from different resolution layers. For instance, in Block1 of PAFPN, the input consists of the up-sampled P5 layer and the sibling P4 layer, which overlooks the importance of shallow low-level spatial information in the P3 layer. Similarly, in Block2, there is no direct fusion of the P2 layer, which contains crucial information about small targets. This deficiency persists in the last two blocks as well.
Secondly, the architecture's strategy for the small target detection layer is formulated through a singular down-top pathway and two blocks, significantly impairing the model's proficiency in learning and representing minute object features. As shown in Fig.~\ref{mafpn}(b) and (c), YOLOv8n based on PAFPN exhibits lower activation capacity in different scale objects compared to the MAFPN proposed in this paper.

The main contributions of this paper are summarized as follows:
\begin{enumerate}[leftmargin=*, labelindent=1em, labelsep=1em]
  \item We propose a new plug-and-play neck called Multi-Branch Auxiliary FPN (MAFPN) to achieve richer feature interaction and fusion. In MAFPN, Superficial Assisted Fusion (SAF) maintains shallow backbone information via bi-directional connectivity, enhancing the network's ability to detect small targets. Additionally, Advanced Assisted Fusion (AAF) enriches the gradient information of the output layer through multi-directional connections. Furthermore, MAFPN can be seamlessly integrated into any other detector to enhance its multi-scale representation capability. 
  \item We designed the lightweight Re-parameterized Heterogeneous Efficient Layer Aggregation Network (RepHELAN) module, which combines the concept of reparameterized heterogeneous large convolutions. This module expands the scope of perception by parallelizing a large kernel convolution with several small kernel convolutions without incurring additional inference costs, while preserving information about small objects.
  \item We propose a Global Heterogeneous Kernel Selection mechanism (GHSK), which adaptively enlarges the effective receptive field of the entire network by adjusting the kernel sizes in RepHELAN across different resolution feature layers in the network architecture.
    \item The Multi-Branch Auxiliary Fusion YOLO (MAF-YOLO) demonstrates superior object detection performance across various aspects on the MS COCO dataset, outperforming existing real-time object detectors.
\end{enumerate}

\section{Related works}
\subsection{Real-time object detectors}
The task of object detection is to identify objects in specific scenes. While both two-stage~\cite{cascade,convnext} and transformer-based~\cite{deformable, dino} detectors achieve high accuracy, their complex structures often entail significant parameter and computational overhead, compromising real-time performance. Most real-time object detection networks employ single-stage methods, with the YOLO series being particularly prominent. PPYOLOE~\cite{ppyoloe} and YOLOv6~\cite{yolov6} explore reparameterization techniques and adopted the Task Alignment Learning (TAL)~\cite{tal} strategy in label assignment, significantly enhancing performance. YOLOv7~\cite{yolov7} proposes the ELAN scheme to optimize the Cross Stage Partial Network structure from YOLOv4~\cite{yolov4} and designs several trainable bag-of-freebies methods for lossless model optimization. YOLOv8~\cite{yolov8} combined and optimized the strategies of current advanced YOLO algorithms to achieve a better balance between speed and accuracy, which is widely used in the industry. The latest YOLOv9~\cite{yolov9} introduced the Generalized Efficient Layer Aggregation Network (GELAN) structure and the concept of Programmable Gradient Information (PGI) to address information bottleneck issues in the network. YOLOv9 boasts a highly efficient parameter utilization, achieving the SOTA of the current YOLO family.
\subsection{Multi-scale features fusion for object detection}
The original idea behind the FPN was to enhance the multi-scale detection capability of the network by incorporating cross-scale connections and information exchange. Significant research has been dedicated to optimizing and extending the FPN structure to improve the efficiency of feature fusion. In YOLOv6-v3~\cite{yolov6}, the Bidirectional concatenation (BIC) mechanism is used to better utilize the backbone shallow information, and DAMO-YOLO~\cite{damo} employs Reparameterized Generalized-FPN (RepGFPN) for richer fusion in both backbone and neck. Gold-YOLO~\cite{goldyolo} introduces an advanced Gather and Distribute (GD) mechanism, which simultaneously integrates local and global information within the neck through convolutional and self-attention operations. These approaches significantly alleviate this issue, but further optimization opportunities remain. 
\section{Methodology}
\subsection{Macroscopic architecture}
\begin{figure}[htb]
	
	\begin{minipage}[b]{1.0\linewidth}
		\centering
		\centerline{\includegraphics[width=11cm]{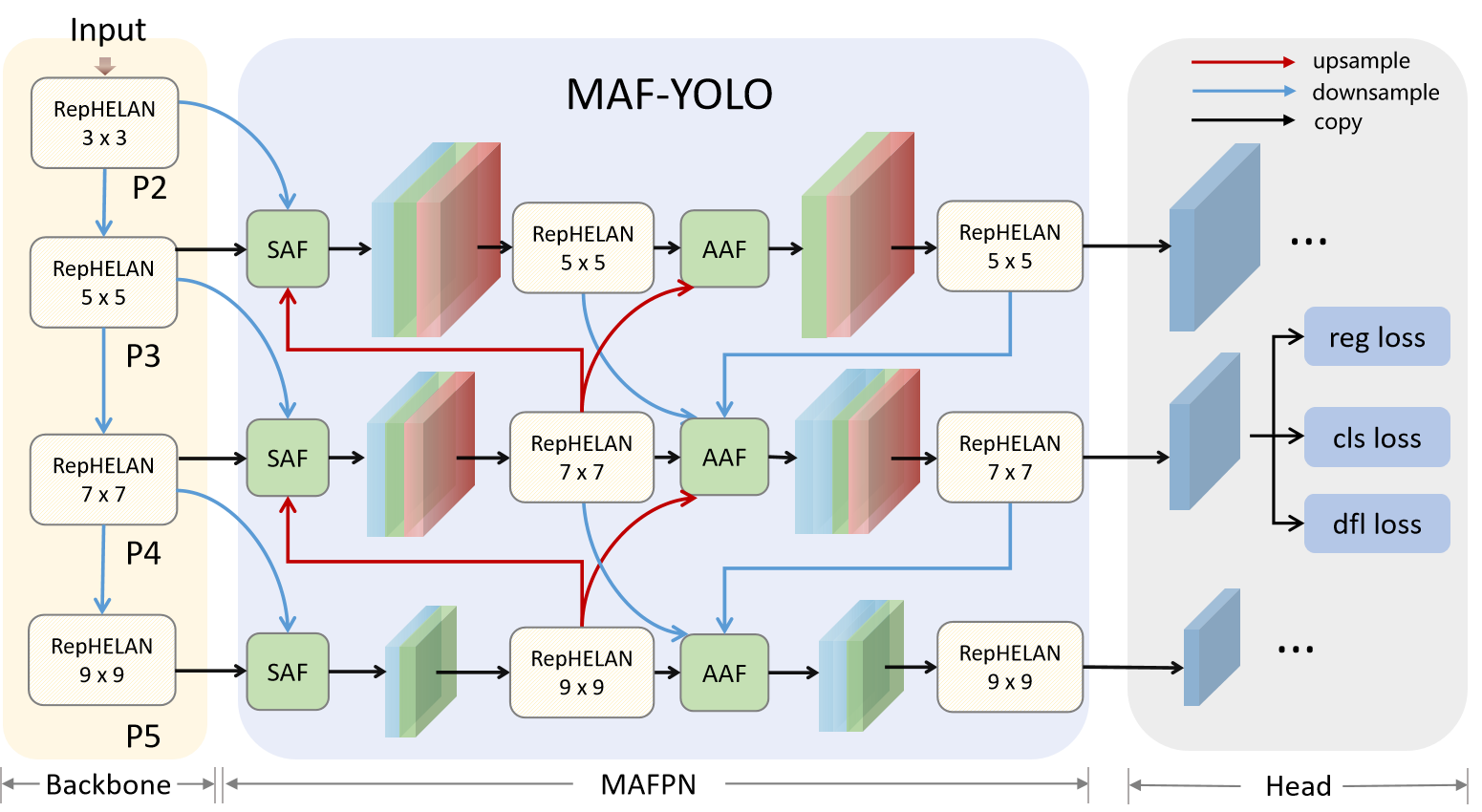}}
	\end{minipage}
	\caption{Overview of the network architecture of MAF-YOLO.}
	\label{mayolo}
\end{figure}

As illustrated in Fig.~\ref{mayolo}, we break down the macroscopic architecture of a one-stage object detector into three main components: the backbone, neck, and head. In MAF-YOLO, the input image initially passes through the backbone, which consists of four stages: P2, P3, P4, and P5. We designed MAFPN as a neck structure. In the first bottom-up pathway, the SAF module is responsible for extracting multi-scale features from the backbone and performing preliminary assisted fusion at the shallow layers of the neck. Meanwhile, AAF collects gradient information from each layer through denser connections in the second top-down pathway, ultimately guiding the head to obtain diversified output information across three resolutions. Both of the aforementioned structures employ the RepHELAN module for feature extraction, which utilizes dynamically sized convolutional kernels to achieve adaptive receptive fields. Finally, the detection heads predict object bounding boxes and their corresponding categories based on feature maps at each scale to compute their loss.
\subsection{Global Heterogeneous Kernel Selection mechanism}
An important factor contributing to the effectiveness of transformers is their self-attention mechanism, which performs query-key-value operations over a global or larger window scale. Similarly, large convolutional kernels capture both local and global features, and the use of moderately large convolutional kernels to increase the effective receptive field has been demonstrated in several works to be effective. Research conducted by Trident Network~\cite{trident} suggests that networks with larger receptive fields are preferable for detecting larger objects, while inversely, smaller-scale targets benefit from smaller receptive fields. YOLO-MS~\cite{yoloms} introduced the concept of Heterogeneous Kernel Selection (HKS) protocols. Employing an incremental convolutional kernel design of 3, 5, 7, and 9 in the backbone to balance performance and speed. Inspired by this work, we extend it to the Global Heterogeneous Kernel Selection (GHKS) mechanism, integrating the concept of heterogeneous large convolutional kernels throughout the entire MAF-YOLO architecture. In addition to the progressively increasing convolutional kernels in RepHELAN of the backbone, we also introduce large convolutional kernels of 5, 7, and 9 in MAFPN to adapt to the requirements of different resolutions, thus progressively obtaining multi-scale sensory field information.

\subsection{Multi-Branch Auxiliary FPN}
Accurate localization relies on detailed edge information from shallow networks, while precise classification requires deeper networks to capture coarse-grained information~\cite{yolov3}. We believe that an effective FPN should support full and sufficient convergence of shallow and deep network information flows.

\begin{figure}[hbb]
	
	\begin{minipage}[b]{1.0\linewidth}
		\centering
		\centerline{\includegraphics[width=6.8cm]{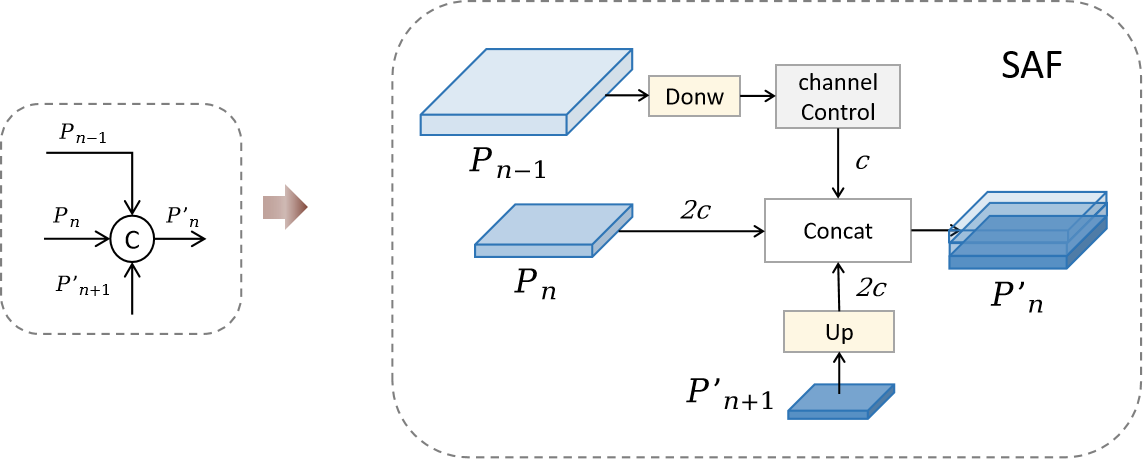}}
	\end{minipage}
	\caption{The architecture of Superficial Assisted Fusion.}
	\label{SAF}
\end{figure}
\subsubsection{Superficial Assisted Fusion.}Preserving shallow spatial information in the backbone is crucial for enhancing the detection capability of smaller objects. However, the information supplied by the backbone is relatively elementary and prone to interference. Therefore, we incorporate shallow information as assisted branches into the deeper network to ensure the stability of subsequent layer learning. Following these principles, we have developed the SAF module, delineated in Fig.~\ref{SAF}. The primary objective of SAF is to integrate deep-level information with features from the same hierarchical level and high-resolution shallow layers within the backbone, aiming to preserve abundant localization details to enhance the spatial representation of the network. Additionally, we utilize $1 \times 1$ convolutions to control the number of channels in shallow layer information, ensuring it occupies a smaller proportion during the concat operation without affecting subsequent learning. Let $P_{n-1}$, $P_n$ and $P_{n+1} \in R^ {H \times W \times C}$ represent the feature maps at different resolutions, where $P_{n}$, $P^{\prime}_{n}$ and $P^{\prime\prime}_{n} $ denote the feature layers of the backbone, and the two paths of the MAFPN. The notation $ U(\cdot) $ signifies the up-sampling operation. $Down$ denotes a $ 3 \times 3$ downsampling convolution accompanied by a batch normalization layer, and $ \delta $ represents a $ silu $ function, $C$ represents the $ 1 \times 1$ convolution of the number of control channels. The output result after applying SAF is as follows:
\begin{equation}
  P^{\prime}_n = concat(\delta(C(Down(P_{n-1}))), P_{n}, U(P^{\prime}_{n+1}))     
\end{equation}

\begin{figure}[htb]
	
	\begin{minipage}[b]{1.0\linewidth}
		\centering
		\centerline{\includegraphics[width=6.8cm]{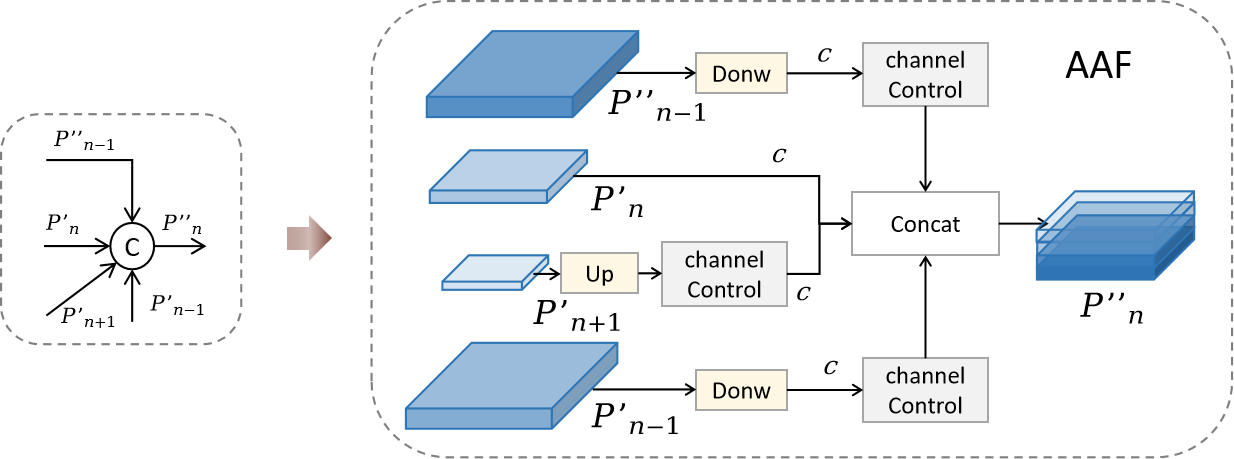}}
	\end{minipage}
	\caption{The architecture of Advanced Assisted Fusion.}
	\label{AAF}
\end{figure}
\subsubsection{Advanced Assisted Fusion.}To further enhance the interactive utilization of feature layer information, we employ the AAF module in the deeper layers of the MAFPN for multi-scale information integration.
Specifically, Fig.~\ref{AAF} illustrates the AAF connections in $P^{\prime\prime}_{n}$, which involve information aggregation across the shallow high-resolution layer $P^{\prime}_{n+1}$, the shallow low-resolution layer $P^{\prime}_{n-1}$, the sibling shallow layer $P^{\prime}_{n}$, and the previous layer $P^{\prime\prime}_{n-1}$.
At this moment, the final output layer P4 can merge information from four different layers simultaneously, thereby significantly enhancing the performance of medium-sized targets. AAF also employs $ 1 \times 1$ convolutional control channels to regulate the impact of each layer on the outcome. Through experimentation, we found that when the strategy in SAF is used, i.e., the number of channels in the three shallow layers is set to half of the number of channels in the deeper layers, which in turn results in a slight degradation of performance. Drawing from the conventional single-path architecture of the FPN, we postulate that the initial guiding information is already embedded within the shallow layers of the MAFPN. Consequently, we equalize the number of channels across each layer to ensure the model obtains diverse outputs. The output result after applying AAF is as follows:
\begin{equation}
  P^{\prime\prime}_n = concat(\!\delta(C(Down(P^{\prime}_{n-1}))), \delta(C(Down(\!P^{\prime\prime}_{n-1}))), P^{\prime}_{n}, C(U(P^{\prime}_{n+1})))    
\end{equation}

\subsection{Re-parameterized Heterogeneous Efficient Layer Aggregation Network}

\begin{figure}[htb]
	
	\begin{minipage}[b]{1.0\linewidth}
		\centering
		\centerline{\includegraphics[width=10cm]{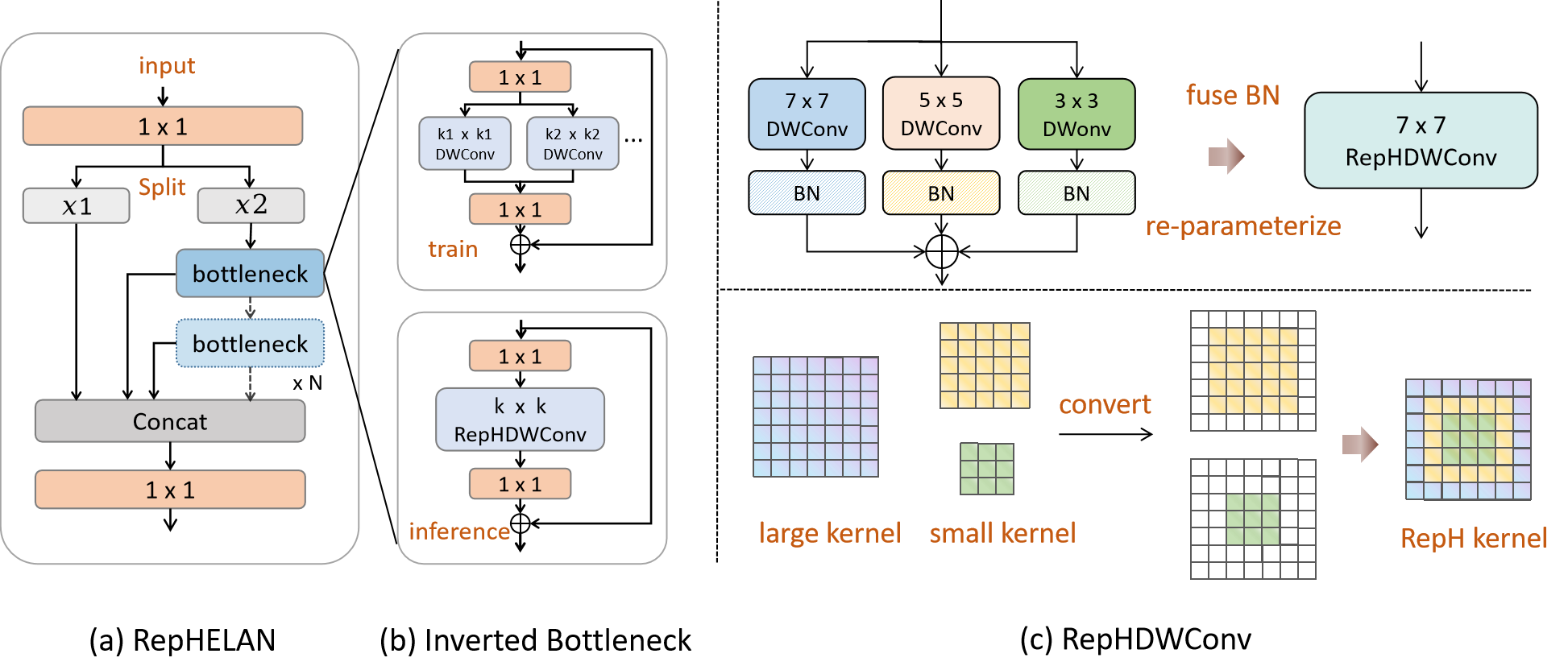}}
	\end{minipage}
	\caption{(a) Overview of the network architecture of RepHELAN, (b) The structure of Inverted Bottleneck in the training and inferencing phases, which is in the RepHELAN, (c) The reparameterization process of a 7×7 RepHDWConv.
}
	\label{RepHELAN}
\end{figure}
After designing the MAFPN structure in the preceding section, another challenge lies in efficiently designing the feature extraction block within the entire architecture. In this section, we present the design of a powerful encoder architecture, which efficiently learns expressive multi-scale feature representations. The structure of RepHELAN is shown in Fig.~\ref{RepHELAN}(a). Initially, the input information undergoes a $1 \times 1$ convolution and a Split operation, resulting in two streams. One stream preserves the original information, which then directly enters the Concat operation, while the other stream undergoes processing through N Inverted Bottleneck units. Due to the mechanism of ELAN, the branches and the outputs passing through each Inverted Bottleneck are retained and eventually concatenated together. The specific structure of the Inverted Bottleneck is illustrated in Fig.~\ref{RepHELAN}(b), where the input sequentially passes through a $ 1 \times 1$ convolution to expand the number of channels, followed by a $ k \times k$ RepHDWConv operation, and finally by $ 1 \times 1$ point convolution to shrink the number of channels and compensate for the possible loss of information caused by DWConv.
\subsubsection{Re-parameterized Heterogeneous Depthwise Convolution}
Firstly, we employed Depthwise convolution with a large kernel in the global architecture to implement the aforementioned GHKS mechanism. Our study also indicates that while larger convolutional kernels may enhance performance by encoding more extensive regions, they might inadvertently obscure details relevant to small targets, thus leaving room for further improvement. Therefore, we transferred the heterogeneous idea from the global architecture to a single convolution and incorporated the idea of Re-parameterization~\cite{repvgg,replknet} to realize RepHConv. Specifically, we complement the detection of small targets by concurrently running large and small convolutional kernels. Different sizes of convolution kernels enhance both the network's ERF and the diverse representation of features. As shown in Fig.~\ref{RepHELAN}(b), the Inverted Bottleneck exhibits certain differences between training and inference. During training, the network runs n parallel depthwise convolutional (DWConv) operations of varying sizes, while during inference, these convolutions are merged into one, resulting in no decrease in inference speed. We believe that RepHDWConv is a superior convolutional strategy that enhances the representation capability across multiple scales with minimal loss.

Steps for reparameterizing a $7 \times 7$ RepHDWConv are shown in Fig.~\ref{RepHELAN}(c). Firstly, a $k_1 \times k_1$ large DWConv and many $k_2 \times k_2$ small DWConv will be parallelized in a RepHDWConv, Each DWConv is followed by a batch normalization (bn) layer and the parameters of each convolution kernel will be merged with the parameters of its corresponding bn layer. In the second step, through a procedure akin to padding, these small DWConvs are assimilated into a larger DWConv, followed by re-parameterization. The parameters and biases of these heterogeneous DWConvs are accumulated to form a new RepHDWConv. Let $I$ indicate input feature maps, $K_{n}$, and $B_{n}$ show the weight and bias of the convolution with a $n \times n$ kernel. The resulting output feature map $O$ is:
\begin{equation}
O = I \otimes \left( K_{2n-1} + \sum_{i=1}^{m} K_{2n-(2i+1)} \right) \\+ \left( B_{2n-1} + \sum_{i=1}^{m} B_{2n-(2i+1)} \right)
\end{equation}
$ \text{where } n \geq 3 \text{ and } m \text{ is the largest integer such that } 2n-(2m+1) \geq 3$. 
\section{Experiments}
\subsection{Experimental Setup}
\subsubsection{Datasets.}We conducted extensive experiments on the Microsoft COCO 2017~\cite{coco} dataset to validate the effectiveness of the proposed MAF-YOLO. Specifically, The training of all methods is conducted on the 115k training images and we report results on the 5000 validation images for the ablation study. We report the results of the standard mean average precision (AP) at various IoU thresholds and target scales.
\subsubsection{Implimentation details.}
Our implementation is based on the YOLOv6-2.0 framework. All experiments are conducted with 8 NVIDIA GeForce RTX 2080Ti GPUs, and all the scales of MAF-YOLO are trained from scratch for 300 epochs without relying on other large-scale datasets, like ImageNet~\cite{imagenet}, or pre-trained weights. In addition to this, we employ stronger dynamic cache-based mixup~\cite{mixup} and mosaic mechanisms and simply replace the two $3 \times 3$ convolutions in the YOLOv6 output header with the lightweight RepHDWConv. More implementation details can be found in the supplementary materials.

\subsection{Analysis of RepHELAN}
In this subsection, we will perform a series of ablation studies on the RepHELAN module. By default, we use the MAF-YOLO nano for all experiments.
\subsubsection{Different computational blocks.}We first do ablation experiments of RepELAN module with various computational blocks from other advanced YOLO models in Tab.~\ref{DifferentBlock}. Our RepHELAN not only has a higher parameter utilization rate compared to other modules but also achieves higher accuracy.

\begin{table}
	\centering
	\caption{The impact of different computational blocks on the MAF-YOLOn.}\label{DifferentBlock}
	\renewcommand\arraystretch{1.0}	
	\setlength{\tabcolsep}{6mm}
		\begin{tabular}{cccc}
		\hline
		  Computational Blocks &  Param & FLOPs & $AP$ \\
    \hline
	      C3~\cite{yolov5} &4.5M &12.9G&41.0\\
	     C2f~\cite{yolov8} &4.7M&13.8G&41.3\\
	      CSPNextBlock~\cite{rtmdet} &3.7M&10.6G &41.2\\
	    \rowcolor{gray! 8} RepHELAN &3.9M&11.1G&42.4\\  
		\hline
		\end{tabular}
\end{table}
\subsubsection{Ablation study on RepHELAN.}
As seen in Tab.~\ref{ABRepHELAN}, we have performed an ablation study on the RepHELAN module, where $LK$ represents whether the idea of large convolution kernels is used in Inverted Bottleneck. Each Bottleneck contains a $ 5 \times 5$ DWConv by default. When using large convolutional kernels, the network follows the GHSK strategy, employing $ 3 \times 3$, $ 5 \times 5$, $ 7 \times 7$, and $ 9 \times 9$ DWConvs across the RepHELAN of the architecture. First, we added the ELAN mechanic, which gives a 0.2\% AP boost and increases the number of counters by a small amount. The third row in the table means that the RepHConv achieves a 0.4\% performance improvement without the added overhead of employing a large convolutional kernel while maintaining the model size unchanged. Furthermore, using only large convolutional kernels and ELAN strategies leads to significant performance gains (+1\% $AP$), albeit with a decrease in performance for small targets (-0.3\% $AP_{s}$). Ultimately, when replacing the large DWConv with RepHConv, we achieved an optimal performance of 42.4\% AP, with noticeable improvements across small, medium, and large object categories.

\begin{table}
	\centering
	\caption{Ablation study on RepHELAN structure.}\label{ABRepHELAN}
	\renewcommand\arraystretch{1.0}	
	\setlength{\tabcolsep}{1mm}
		\begin{tabular}{ccc|cc|ccccc}
		\hline
		ELAN &LK& RepHConv &Param &FLOPs & $AP$ & $AP_{50}$& $AP_{s}$&  $AP_{m}$&  $AP_{l}$\\
    \hline
	      &  & & 3.5M & 10.2G & 40.3& 56.8& 21.6 &45.1 & 55.0\\
            \checkmark&  && 3.6M &  10.4G & 40.5&57.0& 21.8& 45.5&56.1 \\
	    \checkmark&  &\checkmark& 3.6M & 10.4G & 40.9& 57.3& 21.7& 45.6 & 56.4\\
	    \checkmark&  \checkmark & &3.8M&10.5G & 41.5& 58.0& 21.5& 46.0 & 57.5   \\
	    &  \checkmark&\checkmark&3.6M&10.4G & 42.1& 58.2& 21.9& 46.2 & 58.2 \\  
     \rowcolor{gray! 8}\checkmark&  \checkmark&\checkmark&3.8M&10.5G & 42.4& 58.9& 22.0& 46.5 & 59.4\\ 
		\hline
		\end{tabular}
\end{table}

\subsection{Analysis of MAFPN}
In this subsection, we conducted ablation experiments on each module of MAFPN and demonstrated the plug-and-play capability of MAFPN by replacing the neck structure with different algorithms in various experiments.
\subsubsection{Ablation study on MAFPN.}
The results of this experiment, as shown in Tab.~\ref{ABMAFPN}, and the default neck of the model is set to PAFPN,  which includes six RepHELAN Blocks. Firstly, we incorporated SAF modules into the shallow layers of the backbone and neck, which resulted in a 0.3\% performance boost with an increase of 0.3M parameters and it's worth noting that through SAF, we achieved a 1\% improvement in performance for small targets. Secondly, with the sole addition of the AAF module, we observed an enhancement in performance specifically for objects across all scales. Ultimately, the maximum performance of the model was obtained when the combination of SAF and AAF was used.

\begin{table}
	\centering
	\caption{Ablation study on MAFPN structure.}\label{ABMAFPN}
	\renewcommand\arraystretch{1.0}	
	\setlength{\tabcolsep}{2.4mm}
		\begin{tabular}{cc|cc|ccccc}
		\hline
		SAF & AAF &Param &FLOPs & $AP$ & $AP_{50}$& $AP_{s}$&  $AP_{m}$&  $AP_{l}$\\
    \hline
	      &  & 3.1M & 8.75G &41.3& 57.7&21.0&45.4&58.4\\
            \checkmark&  &3.4M &  9.6G & 41.6&58.2& 22.0& 46.1&59.0 \\
	    &  \checkmark&3.6M& 9.8G&42.0&58.6&21.6&46.4&59.2 \\
	    \rowcolor{gray! 8}\checkmark&  \checkmark & 3.8M&10.5G & 42.4& 58.9& 22.0& 46.5 & 59.4    \\
	   
		\hline
		\end{tabular}
\end{table}
\subsubsection{Ablation study on other models.}MAFPN can be used as a plug-and-play module for other models and the results are listed in Tab.~\ref{MAFPNother}. Firstly, we replaced PAFPN with MAFPN in the 
mainstream single-stage detector YOLOv8n and changed the number of channels to keep the model smaller. YOLOv8n-MAFPN uses fewer epochs 
(-200 epochs) and fewer parameters and obtains a 2\% AP improvement, reflecting the excellent performance of MAFPN. What's more, we also
verified the effectiveness of MAFPN using the two-stage detector Cascade MaskRCNN~\cite{cascade} in the instance segmentation task.
\begin{table}
	\centering
	\caption{Performance of MAFPN on other object detectors and different tasks.}\label{MAFPNother}
	\renewcommand\arraystretch{1}	
	\setlength{\tabcolsep}{1.5mm}
		\begin{tabular}{cccccc}
		\hline
		Model &  Neck &  Param  & Bbox $AP$ & Segm $AP$ & epochs\\
    \hline
	     \multirow{3}{*}{MAF-YOLOn} & PAFPN &3.1M &41.3&-&300\\
	    & BIC~\cite{yolov6} &3.5M&41.7&-&300\\
	    &  MAFPN &3.8M& 42.4&-&300\\
     \hline
     \multirow{2}{*}{YOLOv8n} & PAFPN & 3.2M&37.3&-&500\\
	    & MAFPN &3.0M&39.3&-&300\\
    \hline
	    \multirow{2}{*}{Cascade MaskRCNN} & FPN &78M &41.9&35.6& 20\\
	    & MAFPN &83M&42.8&36.3 &20\\
		\hline
		\end{tabular}
\end{table}
\subsection{Ablation study on MAF-YOLO} MAF-YOLO contains MAFPN, RepHELAN module, and GHSK strategy, we performed ablation experiments sequentially and the results are shown in Tab.~\ref{ABMAFYOLO}. We first add the MAFPN structure, which increases the number of 0.5M parameters and improves the performance by 2.1\% AP, then by adding the lightweight RepHELAN module, which reduces the number of parameters by 1.2M, the performance is instead improved by 1.1\% AP, and finally, the GHSK method improves the model accuracy by 1.2\% AP with marginal parameter costs.

\begin{table}[h]
	\centering
	\caption{Ablation study on MAF-YOLOn.}\label{ABMAFYOLO}
	\renewcommand\arraystretch{1.0}	
	\setlength{\tabcolsep}{3mm}
		\begin{tabular}{cccccc}
		\hline
		MAFPN & RepHELAN& GHSK &Param &FLOPs & $AP$ \\
    \hline
	      &  & & 4.3M & 10.9G & 37.7 \\
            \checkmark&  && 4.8M  &  12.1G & 39.8   \\
	    \checkmark& \checkmark && 3.6M & 10.4G & 40.9\\
	    \checkmark&  \checkmark &\checkmark &3.8M &10.5G & 42.4    \\
	   
		\hline
		\end{tabular}
\end{table}

\begin{table*}
	\small
	\centering
	\caption{Comparison with state-of-the-art real-time object detectors. $^\ddag$ represents that self-distillation method is utilized, and * refers to train with pertained models.}\label{SOTA}
	\renewcommand\arraystretch{1.02}
	\setlength{\tabcolsep}{0.8mm}
		\begin{tabular}{ccccccccc}
		\hline
		Methods & $AP$& $AP_{50}$&  $AP^{s}$&  $AP^{m}$&  $AP^{l}$& Params& FLOPs &Epoch\\
		\hline
		YOLOv6n$^\ddag$ ~\cite{yolov6} &   37.5  & 53.1 & 17.8&{\itshape} 41.8&55.1& 4.7M&11.4G&\textbf{300}\\
		YOLOv7t~\cite{yolov7}  &  37.4  & 55.2 & 19.0& 41.8& 52.6& 6.2M
		& 13.7G&\textbf{300}\\
		YOLOv8n~\cite{yolov8}  &   37.3  & 52.6 & 18.5& 41.0& 53.5& \textbf{3.2M}
		& \textbf{8.7G}&500\\
            Gold-YOLOn~\cite{goldyolo}  &   39.9  & 55.9 & 19.1& 44.3& 57.8& 5.6M
		& 12.1G &\textbf{300}\\
            RTMDet-t*~\cite{rtmdet}  &   41.0  & 57.4 & 20.7& 45.3& 58.0& 4.9M
		& 16.2G &\textbf{300}\\
		\rowcolor{gray! 8} MAF-YOLOn  &   \textbf{42.4}  & \textbf{58.9} & \textbf{22.0}& \textbf{46.5}& \textbf{59.4}& 3.8M& 10.5G&\textbf{300}\\
		\hline
    \
		YOLOXs~\cite{yolox} & 40.7&59.6&23.9&45.2&53.8& 9.0M& 26.8G&\textbf{300}\\
		RTMDETs* & 44.6&61.7&24.2&49.2&61.9&8.9M&29.6G&\textbf{300}\\
		YOLOv6s$^\ddag$ &  45.0  &61.8 & 24.3& 50.2& 62.7& 18.5M& 45.3G&\textbf{300}\\
  	YOLOv7s AF &  45.1  &61.8 & 25.7& 50.2& 61.2& 11.0M& 28.1G&\textbf{300}\\
		YOLOv8s &  44.9  & 61.8 & 25.7& 49.9& 61.0&11.2M&28.6G&500\\
        YOLOMS-s~\cite{yoloms} &  46.2  & 63.7 & 26.9& 50.5& 63.0&8.1M&31.2G&\textbf{300}\\
        YOLOv9s~\cite{yolov9} &  46.8  & 63.4 & 26.6& \textbf{56.0}& 64.5&\textbf{7.2M}&26.7G&500\\
		\rowcolor{gray! 8} MAF-YOLOs &   \textbf{47.4}  & \textbf{64.3} & \textbf{27.8}& 51.9& \textbf{64.8}& 8.6M&\textbf{25.5G}&\textbf{300}\\
	    \hline
		YOLOv6m* &  50.0  &66.9 & 30.6& 55.4&67.3& 34.9M& 85.8G&\textbf{300}\\
		YOLOv8m &  50.2  & 67.2 & 32.1& 55.7& 66.5&25.9M&78.9G&500\\
		DAMO-YOLOm$^\ddag$~\cite{damo} &  50.4 & 67.2 &25.9 &50.6 &62.5 &28.2M&\textbf{61.8G}&\textbf{300}\\
        YOLOMS~\cite{yoloms} &  51.0  & \textbf{68.6} & 33.1& 56.1& 66.5&22.2M&80.2G&\textbf{300}\\
		\rowcolor{gray! 15} MAF-YOLOm &   \textbf{51.2}  & 68.5 & \textbf{33.2}& \textbf{56.3}& \textbf{67.5}& \textbf{23.7M}&76.7G&\textbf{300}\\
    \hline
    ConvNeXt-T(Mask R-CNN)~\cite{convnext}& 46.2  &68.1 & 30.1& 49.5& 59.5& 48.1M& 262G&36\\
    Deformable DETR~\cite{deformable} &  46.2  &65.2 & 28.8& 49.2& 61.7& 40.1M& 173G&50\\
    DINO-4scale-R50~\cite{dino} &  50.4  &68.3 & 33.3& 53.7& 64.8& 47.7M& 279G&20\\
    
    \hline

	\end{tabular}
\end{table*}
\subsection{Comparison with State-of-the-Arts}
\begin{figure}[htb]
	
	\begin{minipage}[b]{1.0\linewidth}
		\centering
		\centerline{\includegraphics[width=12cm]{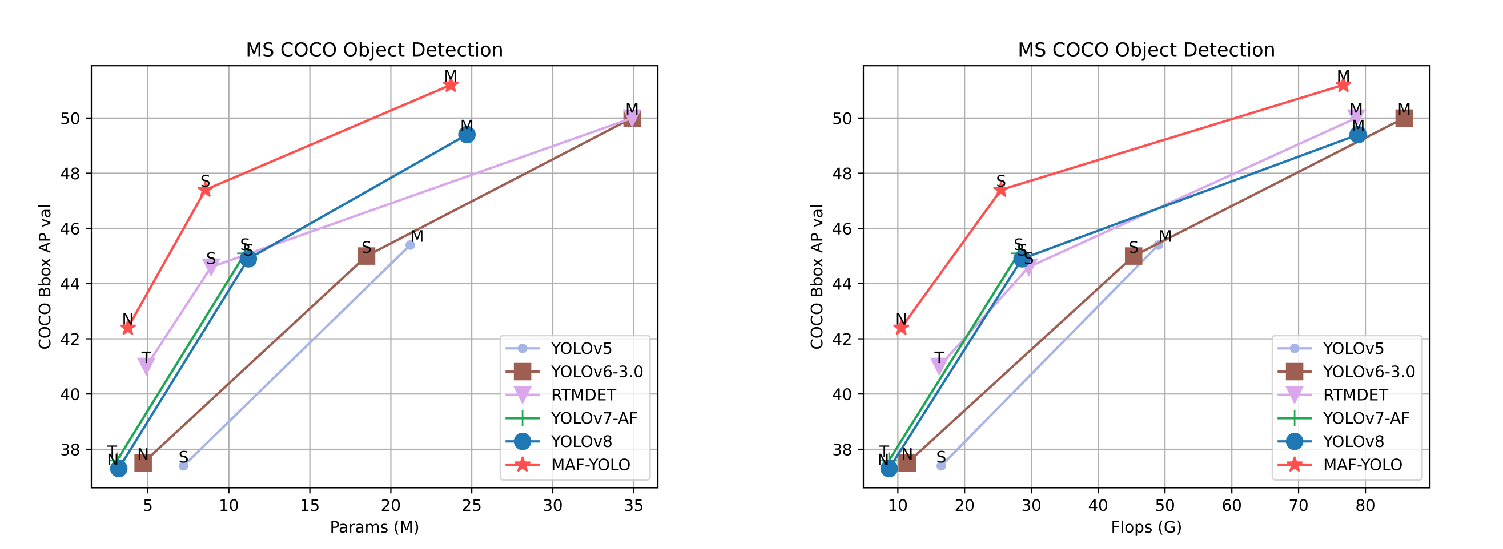}}
	\end{minipage}
	\caption{Comparison of state-of-the-art real-time object detectors.}
	\label{result}
\end{figure}

\begin{figure}[htb]
	
	\begin{minipage}[b]{1.0\linewidth}
		\centering
		\centerline{\includegraphics[width=12cm]{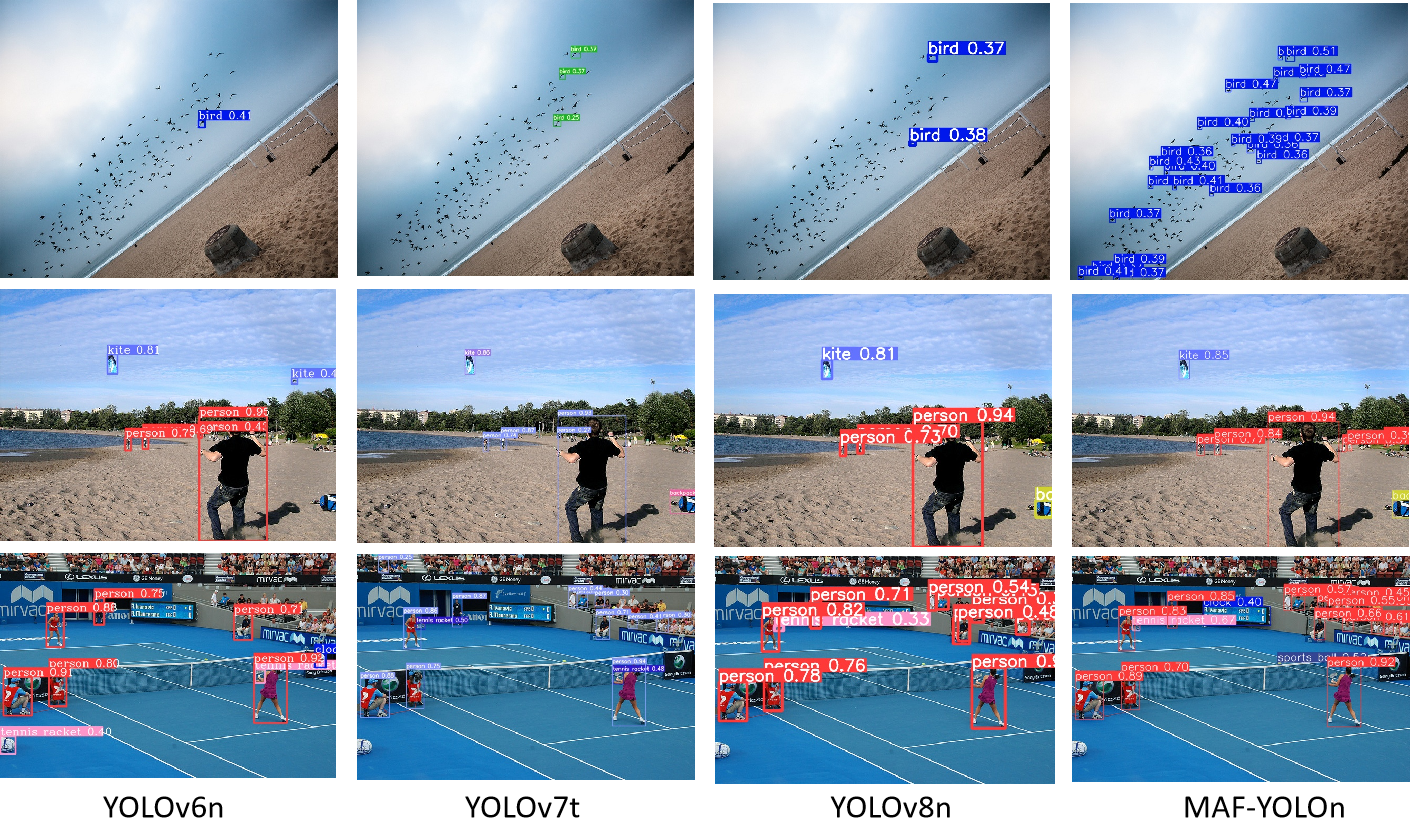}}
	\end{minipage}
	\caption{The detection results of YOLOv6n, YOLOv7t, YOLOv8n, MAF-YOLOn.}
	\label{output}
\end{figure}

Tab.~\ref{SOTA} and Fig.~\ref{result} demonstrate the comparison of our proposed MAF-YOLO with other SOTA real-time target detectors. In comparison to the nano-scale model, MAF-YOLOn has a slightly bigger number of parameters than YOLOv8n, but the AP is improved by 5.1\%. Compared to the current newer Gold-YOLOn, MAF-YOLOn reduces about 36\% of parameters and 13\% of computation but still improves AP by 2.5\%. Our model also has a big advantage for small-scale models, Compared to the anchor-free version of YOLOv7s, MAF-YOLOs has 22\% fewer parameters and has a significant improvement of 2.3\% AP. It is also noteworthy that our MAF-YOLOs achieved comparable results when compared to the current SOTA model YOLOv9s, which is 0.6 AP higher than the YOLOv9s with comparable parameters and calculations. In addition, we present several two-stage and transformer-based detectors where our model demonstrates superior performance and is more lightweight. Some detection results of different algorithms on the COCO validation set are shown in Fig.~\ref{output}.

\section{Conclusions}

In this paper, we introduce MAFPN as a solution to address the limitations of PAFPN in traditional YOLO, which incorporates two key components: SAF and AAF. SAF is employed to effectively retain shallow information in the backbone, while AAF facilitates the output layer in retaining diverse multi-scale information through enhanced information fusion. Furthermore, we integrate GHSK into MAF-YOLO, which dynamically scales up the convolutional kernels throughout the architecture to significantly expand the sensory field of the network. Additionally, we introduce the RepHELAN module, which leverages reparameterized heterogeneous convolutions to greatly enhance the multi-scale characterization capability. As a result, MAF-YOLO demonstrates outstanding overall performance while maintaining a comparable number of parameters.
\subsubsection{Acknowledgements}
This work is supported in part by National Natural Science Foundation of China (72192823, U20A20171, 62373324, 62271448), Natural Science Foundation of Zhejiang Province (LY21F020027, 61972355) and Key Programs for Science and Technology Development of  Zhejiang Province (2022C03113, LZ23F020010).
%
%
%
%
%

\bibliographystyle{splncs04}

\bibliography{icme2023template}

\begin{thebibliography}{10}
\providecommand{\url}[1]{\texttt{#1}}
\providecommand{\urlprefix}{URL }
\providecommand{\doi}[1]{https://doi.org/#1}

\bibitem{yolov4}
Bochkovskiy, A., Wang, C.Y., Liao, H.Y.M.: Yolov4: Optimal speed and accuracy of object detection. arXiv preprint arXiv:2004.10934  (2020)

\bibitem{cascade}
Cai, Z., Vasconcelos, N.: Cascade r-cnn: High quality object detection and instance segmentation. IEEE transactions on pattern analysis and machine intelligence  \textbf{43}(5),  1483--1498 (2019)

\bibitem{yoloms}
Chen, Y., Yuan, X., Wu, R., et~al: Yolo-ms: Rethinking multi-scale representation learning for real-time object detection. arXiv preprint arXiv:2308.05480  (2023)

\bibitem{imagenet}
Deng, J., Dong, W., Socher, R., Li, L.J., Li, K., Fei-Fei, L.: Imagenet: A large-scale hierarchical image database. In: 2009 IEEE conference on computer vision and pattern recognition. pp. 248--255. Ieee (2009)

\bibitem{replknet}
Ding, X., Zhang, X., Han, J., Ding, G.: Scaling up your kernels to 31x31: Revisiting large kernel design in cnns. In: Proceedings of the IEEE/CVF conference on computer vision and pattern recognition. pp. 11963--11975 (2022)

\bibitem{repvgg}
Ding, X., Zhang, X., Ma, N., et~al: Repvgg: Making vgg-style convnets great again. In: Proceedings of the IEEE/CVF conference on computer vision and pattern recognition. pp. 13733--13742 (2021)

\bibitem{tal}
Feng, C., Zhong, Y., Gao, Y., Scott, M.R., Huang, W.: Tood: Task-aligned one-stage object detection. In: 2021 IEEE/CVF International Conference on Computer Vision (ICCV). pp. 3490--3499. IEEE Computer Society (2021)

\bibitem{yolox}
Ge, Z., Liu, S., Wang, F., Li, Z., Sun, J.: Yolox: Exceeding yolo series in 2021. arXiv preprint arXiv:2107.08430  (2021)

\bibitem{yolov8}
Jocher, G., Chaurasia, A., Qiu, J.: Yolo by ultralytics. URL: https://github. com/ultralytics/ultralytics  (2023)

\bibitem{yolov5}
Jocher, G., Chaurasia, A., Stoken, A., et~al: ultralytics/yolov5: v7. 0-yolov5 sota realtime instance segmentation. Zenodo  (2022)

\bibitem{yolov6}
Li, C., Li, L., Geng, Y., et~al: Yolov6 v3. 0: A full-scale reloading. arXiv preprint arXiv:2301.05586  (2023)

\bibitem{trident}
Li, Y., Chen, Y., Wang, N., Zhang, Z.: Scale-aware trident networks for object detection. In: Proceedings of the IEEE/CVF international conference on computer vision. pp. 6054--6063 (2019)

\bibitem{fpn}
Lin, T.Y., Doll{\'a}r, P., Girshick, R., et~al: Feature pyramid networks for object detection. In: Proceedings of the IEEE conference on computer vision and pattern recognition. pp. 2117--2125 (2017)

\bibitem{coco}
Lin, T.Y., Maire, M., Belongie, S., et~al: Microsoft coco: Common objects in context. In: Computer Vision--ECCV 2014: 13th European Conference, Zurich, Switzerland, September 6-12, 2014, Proceedings, Part V 13. pp. 740--755. Springer (2014)

\bibitem{convnext}
Liu, Z., Mao, H., Wu, C.Y., et~al: A convnet for the 2020s. In: Proceedings of the IEEE/CVF conference on computer vision and pattern recognition. pp. 11976--11986 (2022)

\bibitem{rtmdet}
Lyu, C., Zhang, W., Huang, H., et~al: Rtmdet: An empirical study of designing real-time object detectors. arXiv preprint arXiv:2212.07784  (2022)

\bibitem{yolo1}
Redmon, J., Divvala, S., Girshick, R., Farhadi, A.: You only look once: Unified, real-time object detection. In: Proceedings of the IEEE conference on computer vision and pattern recognition. pp. 779--788 (2016)

\bibitem{yolov3}
Redmon, J., Farhadi, A.: Yolov3: An incremental improvement. arXiv preprint arXiv:1804.02767  (2018)

\bibitem{goldyolo}
Wang, C., He, W., Nie, Y., et~al: Gold-yolo: Efficient object detector via gather-and-distribute mechanism. arXiv preprint arXiv:2309.11331  (2023)

\bibitem{yolov7}
Wang, C.Y., Bochkovskiy, A., Liao, H.Y.M.: Yolov7: Trainable bag-of-freebies sets new state-of-the-art for real-time object detectors. In: Proceedings of the IEEE/CVF Conference on Computer Vision and Pattern Recognition. pp. 7464--7475 (2023)

\bibitem{yolov9}
Wang, C.Y., Yeh, I.H., Liao, H.Y.M.: Yolov9: Learning what you want to learn using programmable gradient information. arXiv preprint arXiv:2402.13616  (2024)

\bibitem{pafpn}
Wang, K., Liew, J.H., Zou, Y., Zhou, D., Feng, J.: Panet: Few-shot image semantic segmentation with prototype alignment. In: proceedings of the IEEE/CVF international conference on computer vision. pp. 9197--9206 (2019)

\bibitem{ppyoloe}
Xu, S., Wang, X., Lv, W., Chang, Q., Cui, C., Deng, K., Wang, G., Dang, Q., Wei, S., Du, Y., et~al.: Pp-yoloe: An evolved version of yolo. arXiv preprint arXiv:2203.16250  (2022)

\bibitem{damo}
Xu, X., Jiang, Y., Chen, W., et~al: Damo-yolo: A report on real-time object detection design. arXiv preprint arXiv:2211.15444  (2022)

\bibitem{dino}
Zhang, H., Li, F., Liu, S., et~al: Dino: Detr with improved denoising anchor boxes for end-to-end object detection. arXiv preprint arXiv:2203.03605  (2022)

\bibitem{mixup}
Zhang, H., Cisse, M., Dauphin, Y.N., Lopez-Paz, D.: mixup: Beyond empirical risk minimization. arXiv preprint arXiv:1710.09412  (2017)

\bibitem{deformable}
Zhu, X., Su, W., Lu, L., Li, B., Wang, X., Dai, J.: Deformable detr: Deformable transformers for end-to-end object detection. arXiv preprint arXiv:2010.04159  (2020)

\end{thebibliography}
%




\end{document}